\documentclass[sigplan,nonacm]{acmart}

\usepackage[ruled,linesnumbered,vlined]{algorithm2e}
\usepackage{pifont}
\usepackage{listings}
\usepackage{subcaption}
\usepackage{multirow}
\usepackage{soul}
\usepackage{xspace}
\usepackage{enumitem}
\usepackage{tabularx}
\usepackage{tikz}

\newcommand{\eggmind}{\textsc{EggMind}\xspace}

\newcommand{\enumo}{\textsc{Enumo}\xspace}
\newcommand{\ruler}{\textsc{Ruler}\xspace}

\definecolor{lightgray}{gray}{0.95}
\definecolor{brickred}{rgb}{0.8, 0.25, 0.33}
\definecolor{darkspringgreen}{rgb}{0.09, 0.45, 0.27}

\newcommand{\dhalfcheck}{\textcolor{darkspringgreen}{\ding{51}}\textsuperscript{\textcolor{brickred}{\kern-0.5em\tiny\ding{55}}}}

\soulregister\autoref{7}
\soulregister\ref{7}
\soulregister\cite{7}

\definecolor{LightGray}{gray}{0.97}

\newcommand{\x}{\texttimes\xspace}

\newcommand{\T}[1]{\mbox{\lstinline^#1^}}

\usepackage{xcolor}
\definecolor{codegray}{RGB}{245,245,245}

\usepackage[most]{tcolorbox}

\lstdefinestyle{lispcode}{
  language=Lisp,
  basicstyle=\ttfamily\small,
  columns=fullflexible,
  keepspaces=true,
  showstringspaces=false,
  escapeinside={(*@}{@*)},
  mathescape=true
}

\newtcblisting{graycode}{
  colback=codegray,
  colframe=codegray,
  boxrule=0pt,
  arc=1pt,
  left=6pt,
  right=6pt,
  top=4pt,
  bottom=4pt,
  listing only,
  listing options={style=lispcode}
}

\usepackage{pifont}
\newcommand{\cmark}{\textcolor{green!60!black}{\ding{51}}}
\newcommand{\xmark}{\textcolor{red!75!black}{\ding{55}}}

\usepackage{array}
\usepackage{makecell}

\newcolumntype{C}[1]{>{\centering\arraybackslash}m{#1}}
\AtBeginDocument{%
  }

\setcopyright{none}
 \renewcommand\footnotetextcopyrightpermission[1]{}

\author{Chenyun Yin}
\authornote{Both authors contributed equally to this paper.}
\email{higgs@stu.pku.edu.cn}
\affiliation{%
  \institution{Peking University}
  \city{Beijing}
  \country{China}}
  
\author{Youwei Xiao}
\authornotemark[1]
\email{shallwe@pku.edu.cn}
\affiliation{%
  \department{School of Integrated Circuits}
  \institution{Peking University}
  \city{Beijing}
  \country{China}}

\author{Yuze Luo}
\email{luoyuze@stu.pku.edu.cn}
\affiliation{%
  \institution{Peking University}
  \city{Beijing}
  \country{China}}

\author{Yuyang Zou}
\email{yyzou25@stu.pku.edu.cn}
\affiliation{%
  \institution{Peking University}
  \city{Beijing}
  \country{China}}

\author{Yun Liang}
\authornote{Corresponding author.}
\email{ericlyun@pku.edu.cn}
\affiliation{%
  \department{School of Integrated Circuits}
  \institution{Peking University}
  \city{Beijing}
  \country{China}}

\begin{document}

\title{LLM-Guided Strategy Synthesis for Scalable Equality Saturation}

\begin{abstract}

Equality saturation (EqSat) is a powerful optimization paradigm that compactly represents many equivalent programs in an e-graph and delays commitment until extraction selects a lowest-cost program. Unlike traditional destructive rewriting, EqSat can uncover optimizations that fixed sequential rewrites miss, but in practice often becomes intractable due to rapid e-graph explosion. Making EqSat effective, therefore, requires not only domain-specific rewrite rules but also domain-specific strategies that control how rewrites are organized, how saturation is scheduled, and how the e-graph is simplified during search. Today, much of this strategy design is still manual, making it a major obstacle to automating e-graph-based compilers. Recent rule-synthesis frameworks can automatically infer large rewrite vocabularies from semantic specifications, but they also enlarge the rewrite space and further exacerbate e-graph explosion. Although large language models (LLMs) make automated strategy synthesis plausible, directly evolving backend code remains ineffective in practice. The search lacks reusable strategy abstractions and actionable feedback, and can easily trigger e-graph explosion or converge to poor designs.

We present \eggmind, an LLM-guided, end-to-end framework for synthesizing reusable EqSat strategies. At its core, \eggmind\ introduces a domain-specific language, EqSatL, to represent EqSat strategies as explicit and inspectable artifacts. It then proposes an LLM-guided agentic workflow, equipped with novel techniques including proof-derived rewrite motif caching and tractability guidance, to search efficiently for high-quality strategies while keeping synthesis stable under e-graph growth. Evaluation shows that \eggmind\ substantially improves the resource-quality trade-off on vectorization benchmarks, reducing final cost by 45.1\% and peak RAM by 69.1\% relative to full EqSat. We further show that the same methodology transfers effectively to an XLA-based tensor compiler, and demonstrate its practical potential in a logic-synthesis case study with augmented rewrite spaces.
\end{abstract}

\maketitle

\section{Introduction}
\label{sec:intro}

Equality saturation (EqSat)~\citep{max_willsey_egg_2021,zhang_better_2023} has emerged as a powerful paradigm for program optimization by exploring vast spaces of equivalent programs without premature commitment to specific rewrite sequences. By employing an e-graph to represent an exponential number of variations compactly, EqSat decouples the exploration of equivalent forms from the final extraction based on a cost model. This separation has yielded state-of-the-art results in electronic design automation~\citep{seer_24_chen,emorphic_25_chen,skyegg_25_xiao,Zhang2025ASPENLE, asplos27isamore}, program synthesis~\citep{cao_babble_2022,thomas_automatic_2024,kurashige_cclemma_2024,briggs_megalibm_2024,szalinski2020synthesizing}, and compiler optimization~\citep{vanhattum_vectorization_2021,saiki_target-aware_2024,shaikhha_sparse_diff_2024,zhang_user_sched_tensor_2026,hartmann_tensor_eqsat_mcts_2024}. However, the search space often expands so rapidly that exponential growth in memory and runtime becomes the primary bottleneck before a near-optimal solution can be extracted.

Mitigating this explosion necessitates a focus on \emph{strategy}. In this paper, a \emph{strategy} is a control structure for organizing search, including partitioning rulesets, scheduling saturation, simplifying e-graph states, and budget concerns. Strategy-guided EqSat bridges the gap between exhaustive exploration and rigid rewrites. It preserves the non-destructive nature of equality saturation while constraining search through orchestrated control over the process.
Recent studies~\citep{thomas_automatic_2024,he2023mctsgebmontecarlotree,koehler_guided_eqsat_2024,asplos27isamore} confirm that these strategic choices are the primary determinants of both scalability and the quality of the resulting optimizations.

Despite its importance, manual strategy design remains the primary obstacle to achieving full automation in e-graph-based compilers~\citep{thomas_automatic_2024,koehler_guided_eqsat_2024}. Although rule synthesis has been successfully automated by frameworks like \enumo~\citep{pal_equality_2023} and \ruler~\citep{nandi_rewrite_2021}, these tools address only one aspect of the automation puzzle. Crucially, the resulting influx of rules paradoxically exacerbates e-graph explosion by expanding the search branching factor. Consequently, the focus must shift to \emph{automated strategy design} for determining the specific timing and order of rule applications. Current strategies are largely manual and fragile, relying on static domain-specific assumptions that scale poorly with massive rulesets and fail to generalize across different domains.

While recent progress~\citep{novikov2025alphaevolvecodingagentscientific,chen2026magellanautonomousdiscoverynovel,qiu2026passbypassoptimizationintentdrivenir,mikek2026agenticcodeoptimizationcompilerllm} in LLM-guided code generation suggests a path toward automating this process, directly evolving EqSat strategies remains exceptionally difficult. The primary hurdle is that high-level strategic intent must be explicitly written in low-level, backend-specific languages such as egglog~\cite{zhang_better_2023}, making reusable strategy patterns difficult to expose and use. Furthermore, the feedback loop for LLMs is fundamentally broken, as coarse metrics like final cost offer little insight into \emph{why} a strategy succeeded, while raw e-graph states are far too massive to serve as useful context for the LLM. Moreover, without a way to guide and evaluate the search efficiently, LLMs frequently propose unstable strategies that trigger catastrophic e-graph growth and exhaust resources, or converge to poor solutions.

To overcome these challenges, we present \eggmind, a framework that reformulates strategy synthesis as a controlled offline synthesis process. Our approach rests on three pillars. First, it introduces \emph{EqSatL}, a domain-specific language that reifies EqSat control into inspectable and reusable artifacts, separating strategic intent from backend implementation. Second, it distills successful optimization experiences from representative cases into compact evidence through \emph{proof-derived rewrite motif caching}. Third, it incorporates \emph{tractability guidance} to steer the synthesis toward stable search regimes. Together, these techniques allow \eggmind to transform open-ended low-level code evolution into a structured search for reusable strategy artifacts that can be validated and refined across an entire problem family.

This work offers the following contributions:

\begin{itemize}[leftmargin=*]
    \item We present \eggmind, an {\color{blue}{open-source}}, LLM-guided framework for synthesizing reusable EqSat strategies.
    \item We introduce EqSatL, a DSL that reifies reusable EqSat control as an explicit artifact boundary for ruleset organization, schedule construction, and e-graph simplification.
    \item We formulate an agentic workflow for EqSat strategy synthesis, supported by proof-derived rewrite motif caching and tractability guidance, to distill reusable evidence from successful runs and steer search toward stable, high-quality strategies under e-graph growth.
\end{itemize}

\paragraph{Evaluation.}
We evaluate \eggmind on two benchmark families, vectorization~\citep{thomas_automatic_2024} and XLA-based tensor graph optimization~\citep{xla_tf2xla}, as well as a case study against the EqMap~\cite{hofmann_eqmap_2025} logic synthesis tool.
On the vectorization benchmark family, \eggmind substantially improves the quality--resource trade-offs: it reduces final cost by 45.1\% and peak memory usage by 69.1\% compared to full EqSat, while achieving a 2.21\x geometric-mean runtime speedup over the expert-tuned Isaria strategy.
On the XLA-based benchmark, \eggmind consistently matches or outperforms full EqSat across all 17 cases, achieving lower final cost in 13 instances while delivering an 11.89\x geomean speedup in runtime.
In the EqMap case study, \eggmind achieves the lowest cost across all carry-chain benchmarks, reducing cost by 33.76\% over the standard EqMap baseline and by 8.75\% over a higher-budget baseline, while also reducing peak memory by 51.94\% relative to unguided search in the augmented rewrite space.
These results show that LLM-guided strategy synthesis can improve EqSat quality and efficiency, transfer effectively across different optimization domains, and make augmented rewrite spaces practically exploitable.

\section{Background and Motivation}
\label{sec:background}

This section reviews the background of equality saturation and strategy-guided search, situates our work within the landscape of existing EqSat strategy approaches, and motivates \eggmind by identifying the limitations of direct code evolution. 

\subsection{Equality Saturation and Strategy}

\begin{figure*}[t]
  \centering
  \includegraphics[width=0.9\textwidth]{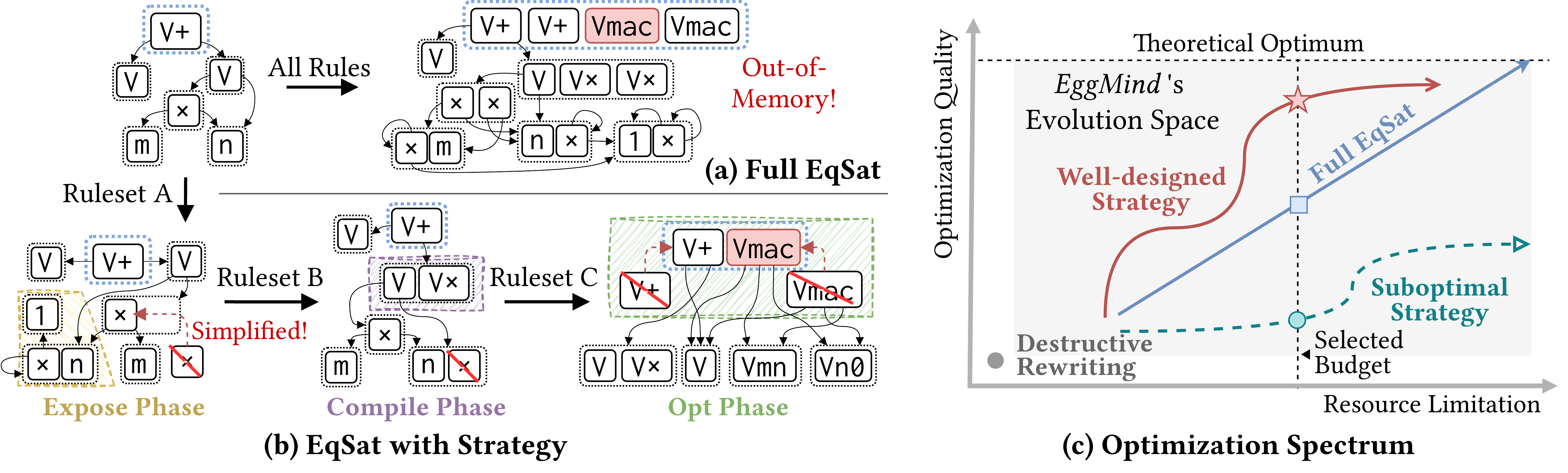}
  \caption{An example comparing (a) full EqSat and (b) strategic EqSat, and (c) the spectrum for their resource--quality trade-off.}
  \label{fig:eqsat-with-strategy}
\end{figure*}

Equality saturation (EqSat) utilizes e-graphs to compactly represent equivalent programs while deferring commitment until extraction selects a lowest-cost program, and is implemented in systems such as \texttt{egg}~\citep{max_willsey_egg_2021} and \texttt{egglog} ~\citep{zhang_better_2023}. The e-graph is organized into \emph{e-classes}, which group equivalent \emph{e-nodes}. Each e-node represents functional operators with other e-classes as their children. As illustrated in \autoref{fig:eqsat-with-strategy}(a), in a vectorization task, EqSat can accumulate diverse equivalent variants, including both scalar and vectorized forms, within a shared e-graph through repeated rewrite applications. This non-destructive search avoids the premature commitment of destructive rewriting. 

In practice, exposing all rewrites at once can trigger rapid e-graph growth and quickly make unguided EqSat intractable under realistic resource budgets~\citep{thomas_automatic_2024, koehler_guided_eqsat_2024, misaal_2025synthesis-based_2025}. To address this problem, a \emph{strategy} organizes the search space without altering rewrite semantics~\citep{thomas_automatic_2024,koehler_guided_eqsat_2024}. It partitions rules into sequenced phases, inserts simplification points, and assigns execution budgets. As shown in \autoref{fig:eqsat-with-strategy}(b), rather than exposing all rules simultaneously, a strategy stages the search: an \texttt{expose} phase first explores scalar structures, a \texttt{compile} phase then lifts them into vector forms, and a final \texttt{opt} phase targets high-profit combinations like \texttt{vecmac}. Intermediate \emph{simplification} steps contract the e-graph between these phases, pruning redundant e-nodes while preserving productive interactions for subsequent stages. In this way, strategy-guided EqSat can contain e-graph growth more effectively than unguided search while retaining the benefits of non-destructive exploration.

\autoref{fig:eqsat-with-strategy}(c) summarizes the resulting resource-quality trade-off. Destructive rewriting is cheap but myopic because it commits too early to a single rewrite path. Full EqSat offers the highest theoretical optimization quality, but in realistic settings, it often exceeds resource constraints before achieving optimality. Strategy-guided EqSat occupies the middle ground: a well-designed strategy can recover much of EqSat's optimization quality under limited budgets, while a suboptimal strategy may over-suppress e-graph growth but underperform even full EqSat. \eggmind is designed to automate this strategy-design process by synthesizing reusable strategies that shift this trade-off toward a more favorable operating point across homogeneous workloads.

\begin{table}[t]
  \centering
  \caption{Comparison of strategy-guided EqSat systems.}
  \label{tab:strategy-landscape}
  \footnotesize
  \setlength{\tabcolsep}{2pt}
  \renewcommand{\arraystretch}{1.0}
  \begin{tabular}{|l|c|c|c|c|}
  \hline
  \textbf{Work} &
  \textbf{Scope} &
  \textbf{Auto?} &
  \textbf{LLM-Guided?} &
  \textbf{Reusable?} \\
  \hline

  \begin{tabular}[c]{@{}l@{}}\textbf{Isaria}~\citep{thomas_automatic_2024}\end{tabular} &
  DSP &
  \xmark &
  \xmark &
  \cmark \\
  \hline

  \begin{tabular}[c]{@{}l@{}}\textbf{MCTS-based}~\citep{he2023mctsgebmontecarlotree, hartmann_tensor_eqsat_mcts_2024}\end{tabular} &
  general &
  \cmark &
  \xmark &
  \xmark \\
  \hline

  \begin{tabular}[c]{@{}l@{}}\textbf{Guide-based}~\citep{koehler_sketch_guided_eqsat_2022,koehler_guided_eqsat_2024}\end{tabular} &
  general &
  \xmark &
  \xmark &
  \xmark \\
  \hline

  \begin{tabular}[c]{@{}l@{}}\textbf{ASPEN}~\citep{Zhang2025ASPENLE}\end{tabular} &
  RTL &
  \cmark &
  \cmark &
  \xmark \\
  \hline

  \begin{tabular}[c]{@{}l@{}}\textbf{\eggmind}~(this work)\end{tabular} &
  general &
  \cmark &
  \cmark &
  \cmark \\
  \hline
  \end{tabular}
\end{table}

\subsection{Strategy Landscape and Open Gaps}

As summarized in \autoref{tab:strategy-landscape}, the current landscape of strategy-guided EqSat systems reveals a fragmented design space.
While the importance of strategic control is well-established~\citep{thomas_automatic_2024,koehler_sketch_guided_eqsat_2022,koehler_guided_eqsat_2024,he2023mctsgebmontecarlotree}, existing approaches prioritize different trade-offs.
Expert-tuned systems like Isaria~\citep{thomas_automatic_2024} offer reusable schedules but rely on labor-intensive manual design.
Guide-based methods~\citep{koehler_sketch_guided_eqsat_2022,koehler_guided_eqsat_2024} effectively steer the search through externally specified signals, yet they stop short of synthesizing reusable strategy artifacts.
Conversely, automated approaches such as MCTS-based search~\citep{he2023mctsgebmontecarlotree,hartmann_tensor_eqsat_mcts_2024} and ASPEN~\citep{Zhang2025ASPENLE} act online during optimization, incurring per-instance overhead without producing reusable strategies.

A critical need persists for a framework that combines fully automated, offline synthesis with cross-case reusability.
This need is intensified by the maturation of automated rule-synthesis tools like \enumo\citep{pal_equality_2023} and \ruler\citep{nandi_rewrite_2021}.
As these systems broaden rewrite vocabularies, the primary bottleneck in compiler construction shifts from rule discovery to strategy synthesis, as larger rewrite spaces demand more sophisticated control to prevent e-graph explosion.
\eggmind addresses this need by synthesizing reusable EqSat strategies from representative workloads, overcoming the limitations of manual scheduling, external guidance, and instance-specific online control.

\subsection{Motivating LLM-Guided Strategy Synthesis}
\label{sec:code-evolution}
\begin{figure}[t]
  \centering
  \includegraphics[width=\columnwidth]{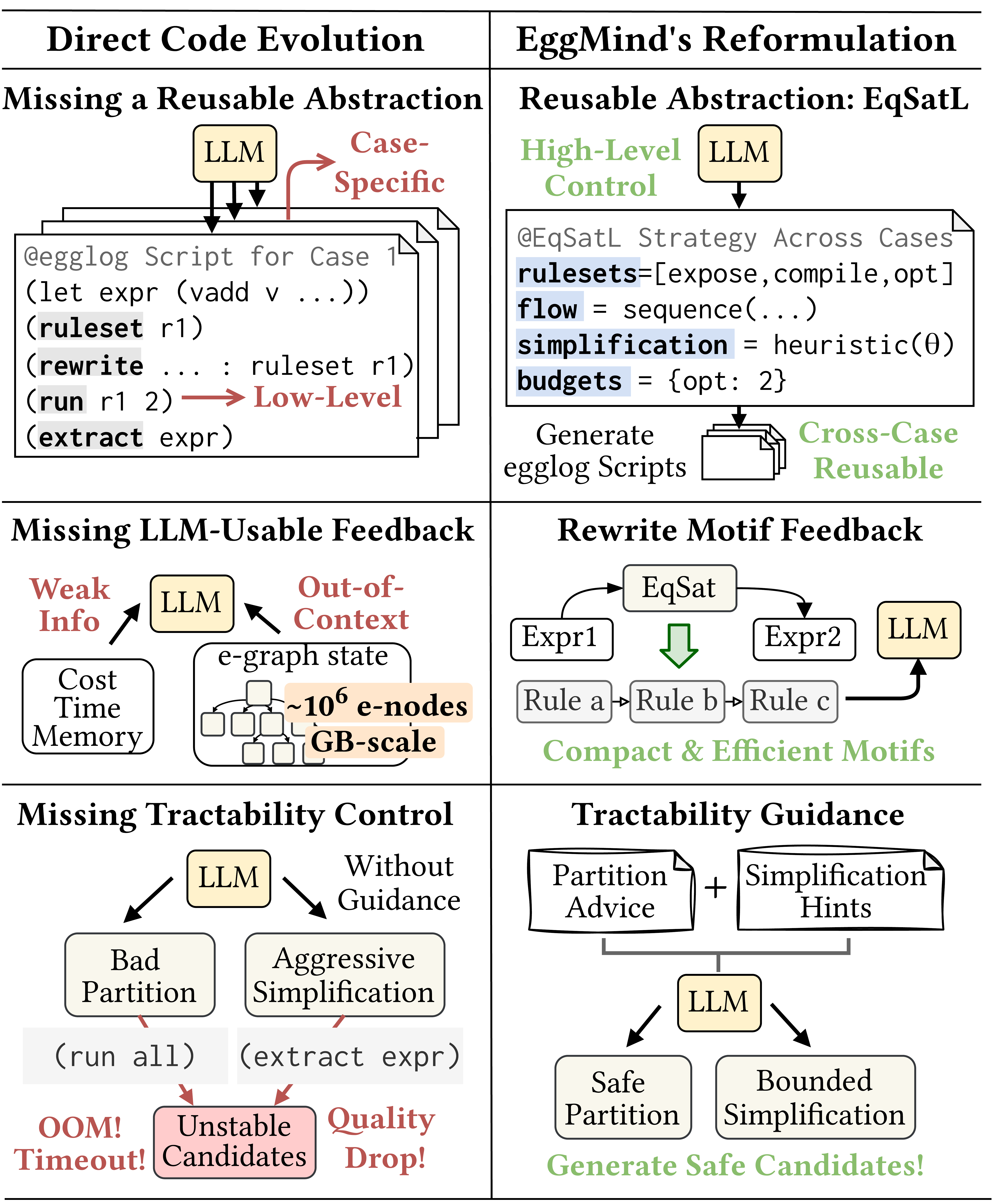}
  \caption{Challenges and \eggmind's solutions}
  \label{fig:code-evolution-challenges}
  \Description{A single-column motivation figure illustrating three difficulties of direct LLM-guided EqSat strategy evolution: no explicit strategy artifact, delayed and expensive feedback, and an unsafe search space that can cause e-graph blow-up, timeouts, or unstable search.}
\end{figure}


Recent advances~\citep{novikov2025alphaevolvecodingagentscientific,chen2026magellanautonomousdiscoverynovel,qiu2026passbypassoptimizationintentdrivenir,mikek2026agenticcodeoptimizationcompilerllm} show that LLMs can effectively guide code generation and algorithm evolution through iterative proposal–feedback loops. Systems such as AlphaEvolve~\citep{novikov2025alphaevolvecodingagentscientific} illustrate this capability in algorithmic discovery, while Magellan~\citep{chen2026magellanautonomousdiscoverynovel} adopts a similar paradigm for EqSat extraction. In contrast to these approaches, which focus on evolving a single heuristic, our goal is to synthesize reusable strategies that govern the entire EqSat workflow across related cases. Under this broader setting, however, directly applying the code-evolution paradigm introduces additional challenges. \autoref{fig:code-evolution-challenges} summarizes these challenges and demonstrates how \eggmind reformulates the problem. The left column highlights three obstacles:

\emph{Missing a reusable abstraction.} Generating rewrite schedule directly towards specific backends yields one-off per-instance scripts that expose only low-level primitives such as \texttt{let}, \texttt{run}, \texttt{extract}, and \texttt{rewrite}. As a result, this mechanism falls short in representing higher-level strategy semantics, such as ruleset organization, phase structure, and simplification control, and reusing them across cases.

\emph{Missing LLM-usable feedback.} Iterative strategy evolution depends on feedback that helps models understand why a candidate succeeds or fails. In EqSat, however, the most obvious signals do not fit that role well: end-to-end metrics such as cost, time, and memory are too weak to explain this, while full e-graph states are too large and unstable to serve as useful model context. Direct interaction, therefore, gives the model either too little information or too much state.

\emph{Missing tractability control.} EqSat strategy synthesis necessitates rigorous control over e-graph expansion. Without guidance, however, LLMs frequently fall back on unstable strategy candidates characterized by bad partitions that trigger unconstrained growth (e.g., \texttt{run all}) or aggressive simplification (e.g., directly extracting the e-graph) that prematurely prunes the e-graph, degrading solution quality.

The right column of \autoref{fig:code-evolution-challenges} shows how \eggmind addresses these three problems. EqSatL provides a reusable abstraction that not only exposes high-level structure directly, but can also be lowered by the backend into multiple per-case egglog scripts, enabling cross-case reuse. Instead of relying on raw execution traces or full e-graph state, \eggmind surfaces compact and efficient feedback by extracting rewrite motifs from egglog proof objects~\citep{zhang_better_2023,flatt_small_proofs_2022,oe_smt_proof_checking_2009,moura_z3_proofs_2008} that connect successful EqSat inputs and outputs through concise rule chains. Finally, it leverages \emph{tractability guidance} to steer synthesis away from unstable partitions and unsafe pruning choices through partition advice and simplification hints. In this way, open-ended code evolution becomes controlled offline strategy synthesis over an inspectable object that can be validated, refined, and reused across related cases.

\section{Overview of \eggmind}
\label{sec:overview}

\begin{figure}[t]
  \centering
  \includegraphics[width=0.99\columnwidth]{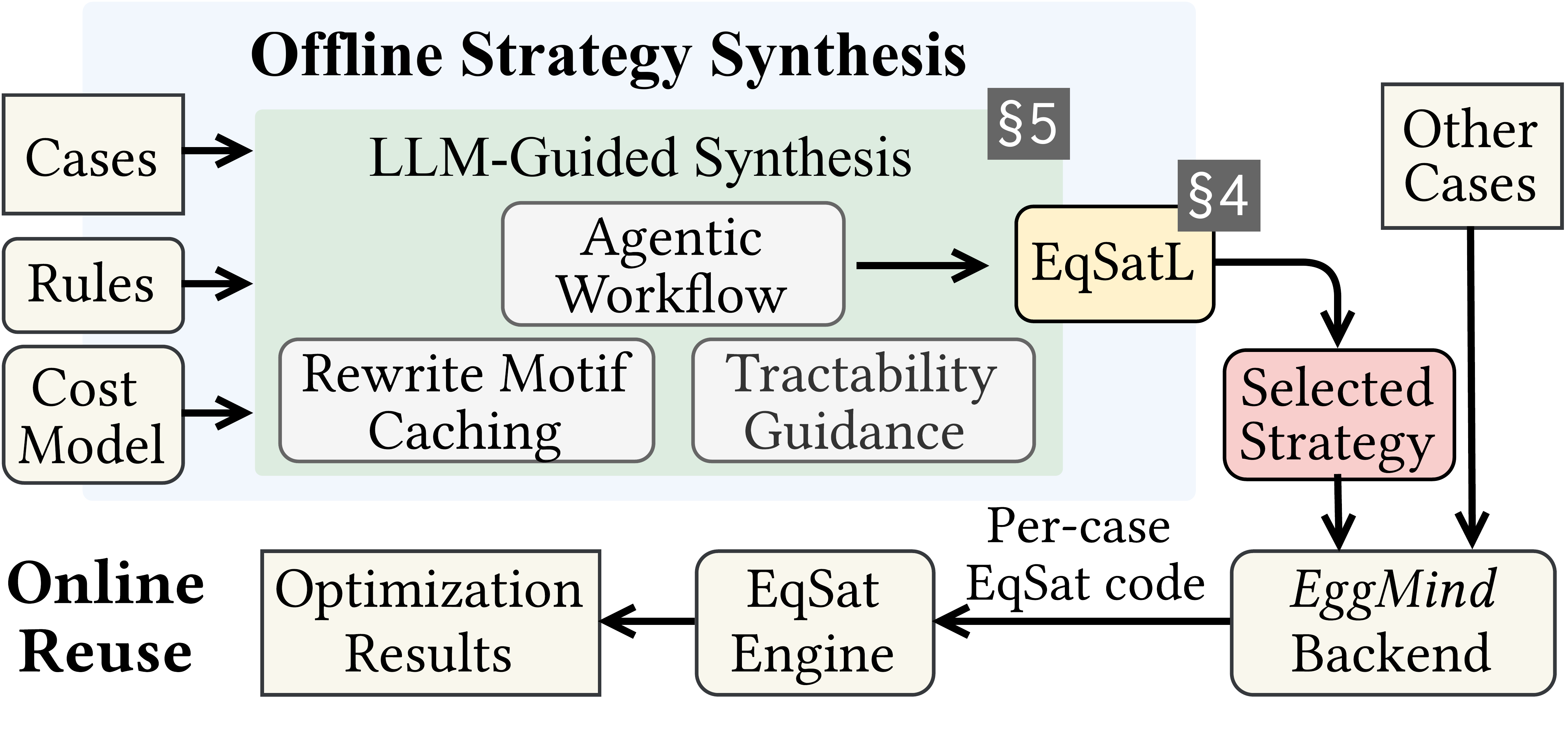}
  \caption{Overview of \eggmind.}
  \label{fig:overview}
\end{figure}

\autoref{fig:overview} gives an overview of \eggmind. The system takes three primary inputs: a collection of cases for evolution, a set of rewrite rules, and a target cost model. Its goal is to synthesize an explicit and reusable EqSat strategy, represented as an EqSatL artifact (\autoref{sec:eqsatl}), for a targeted optimization domain. \eggmind separates this process into two stages: \textit{offline synthesis} and \textit{online reuse}. 
During the offline stage (inside the blue box), \eggmind\ performs \emph{LLM-guided strategy synthesis} (\autoref{sec:guided-synthesis}) to produce EqSatL strategy artifacts. This synthesis process is carried out through the \emph{Agentic Workflow} (\autoref{sec:agent-workflow}) and supported by two additional components: proof-derived rewrite motif caching (\autoref{sec:trace-memory}), which accumulates reusable motifs from successful runs, and tractability guidance (\autoref{sec:tractability-guidance}), which helps keep the search stable in the presence of e-graph growth. The outcome of this offline stage is a selected EqSat strategy artifact for later reuse.
In the online stage (outside the blue box), the selected strategy is combined with other cases and consumed by the \eggmind backend, which lowers the artifact into per-case executable EqSat code. The generated code is then executed by the EqSat engine to produce optimization results. This separation allows the cost of strategy discovery to be paid offline and amortized across future optimization runs. The remainder of the paper first formalizes EqSatL and then presents the agentic workflow together with its supporting memory and tractability-guidance mechanisms.

\section{EqSatL: A DSL for EqSat Strategies}
\label{sec:eqsatl}

\newsavebox{\eqsatlsyntaxbox}
\newlength{\eqsatlsyntaxheight}
\sbox{\eqsatlsyntaxbox}{%
  \small
  \setlength{\tabcolsep}{3pt}
  \begin{tabular}{@{}rcl@{}}
    \multicolumn{3}{@{}l@{}}{\textbf{Artifact}} \\[2pt]
    $\mathit{strategy}$ & $::=$ & $\textsf{strategy}(name,\; \mathit{rulesets},\; \mathit{flow})$ \\[6pt]

    \multicolumn{3}{@{}l@{}}{\textbf{Ruleset Partitioning}} \\[2pt]
    $\mathit{rulesets}$ & $::=$ & $\mathit{ruleset}^{+}$ \\
    $\mathit{ruleset}$ & $::=$ &
    $\textsf{ruleset}(name,\; \textsf{tags}=[\mathit{tag}^{*}])$ \\[6pt]

    \multicolumn{3}{@{}l@{}}{\textbf{Schedule Construction}} \\[2pt]
    $\mathit{flow}$ & $::=$ & $\mathit{node}$ \\
    & $\mid$ & $\textsf{sequence}(\mathit{node}^{+})$ \\
    $\mathit{node}$ & $::=$ &
    $\textsf{phase}(name,\; ruleset,\; iter\_limit)$ \\
    & $\mid$ &
    $\textsf{repeat}(\mathit{flow},\; rounds)$ \\
    & $\mid$ &
    $\mathit{simplify}$ \\[6pt]

    \multicolumn{3}{@{}l@{}}{\textbf{Simplification Control}} \\[2pt]
    $\mathit{simplify}$ & $::=$ &
    $\textsf{heuristic\_simplify}(\theta)$ \\
    & $\mid$ &
    $\textsf{hint\_guided\_simplify}(\theta,\; \mathit{hints})$ \\
    $\mathit{hints}$ & $::=$ &
    $\mathit{hint\_pair}^{*}$ \\
    $\mathit{hint\_pair}$ & $::=$ &
    $\left(\mathit{preferred\_pattern},\; \mathit{pruned\_pattern}\right)$
  \end{tabular}%
}
\setlength{\eqsatlsyntaxheight}{\dimexpr\ht\eqsatlsyntaxbox+\dp\eqsatlsyntaxbox\relax}

\begin{figure*}[t]
  \centering
  \begin{minipage}[b]{0.35\textwidth}
    \centering
    \subcaptionbox{Formal syntax of EqSatL.\label{fig:eqsatL-syntax}}{
      \resizebox{\linewidth}{!}{\usebox{\eqsatlsyntaxbox}}%
    }
  \end{minipage}
  \hspace{1.5em}
  \begin{minipage}[b]{0.52\textwidth}
    \centering
    \subcaptionbox{A compact EqSatL view of an Isaria reference strategy.\label{fig:eqsatL-example}}{
      \includegraphics[width=\linewidth]{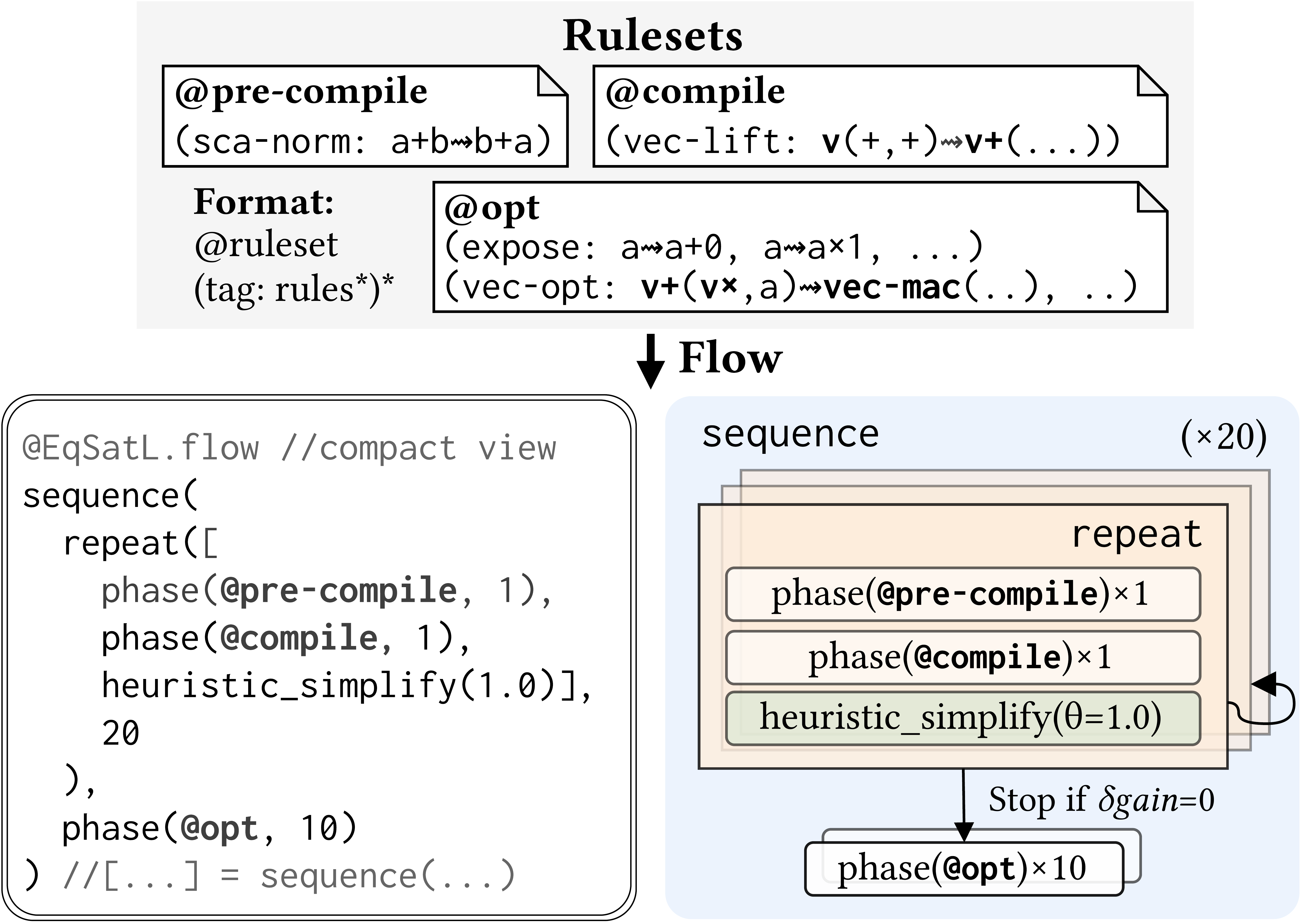}
    }
  \end{minipage}

  \caption{EqSatL: (a) formal syntax and (b) an example strategy.}
  \label{fig:eqsatL}
\end{figure*}
EqSatL is a domain-specific language (DSL) that represents EqSat strategies as explicit artifacts. 
By decoupling strategic intent from concrete execution, it acts as an orchestration layer over backend engines such as egglog~\citep{zhang_better_2023}, leaving per-case rewrite execution to the backend while preserving the strategy structure needed for cross-case synthesis and reuse.
Its design must therefore balance three competing requirements: compactness for stable LLM generation, explicitness for formal validation, and expressiveness for capturing the full optimization potential of EqSat. 
Our formulation of the DSL is informed by two key observations about the trade-off between optimization opportunity and e-graph tractability. \emph{Rewrite interactions} enable high-profit transformations, but they can also trigger the e-graph explosion characteristic of unguided search. \emph{State pruning} serves as the essential counterweight. Without it, redundant states accumulate and eventually exhaust the search budget. According to these two observations, we cast strategy design into three control questions: which rewrites may interact, how their interaction is organized over time, and how the e-graph is contracted afterward. EqSatL follows this decomposition through three control surfaces, namely ruleset partitioning, schedule construction, and simplification control, as shown in \autoref{fig:eqsatL-syntax}.

\paragraph{Ruleset Partitioning.}
Ruleset partitioning determines which rewrites may interact during EqSat search. In EqSatL, the rewrite vocabulary is partitioned into rulesets through semantic \emph{tags} rather than plain rule lists, giving the synthesis process a more compact vocabulary for strategy construction. These tags are defined once per domain by asking the LLM to group rewrites into fine-grained semantic categories, and are then reused throughout strategy synthesis.

\paragraph{Schedule Construction.}
Schedule construction governs how ruleset interaction unfolds over time. EqSatL organizes execution as a \emph{flow tree} that composes ruleset applications into ordered phases, sequences, and \texttt{repeat} regions. Phases and sequences define a linear search plan, while \texttt{repeat} lets a bundle of phases be revisited as a single interaction unit. 
Each iteration ends with a mandatory \texttt{simplify} step, which bounds e-graph growth while still allowing multiple rulesets to react to one another repeatedly and expose optimization opportunities that local saturation would miss. This gives EqSatL a structured way to manage search depth and cross-phase interaction without triggering the instability often caused by unconstrained looping.

\paragraph{Simplification Control.}
Even when rewrite interaction is well structured through partitioning and scheduling, it can still cause substantial e-graph growth. EqSatL therefore reifies \texttt{simplify} as a first-class control node rather than leaving it implicit. By default, simplification is heuristic, with aggressiveness controlled by a strength parameter $\theta$. In our implementation, this is realized either by extract-and-rebuild contraction or by within-eclass cost-based pruning. EqSatL also supports \emph{LLM-hinted simplification}, which uses optional LLM-proposed hint pairs to express structural preferences about which forms should be prioritized or pruned more aggressively in a given context. Simplification thus becomes an explicit strategic decision about when and how to contract the e-graph. The detailed implementation of LLM-hinted simplification is discussed in \autoref{sec:tractability-guidance}.

\autoref{fig:eqsatL-example} illustrates a representative EqSatL strategy in two parts. The upper panel shows the ruleset partition using representative semantic tags and example rewrites (full set omitted for brevity): \texttt{pre-compile} groups scalar normalization rewrites, \texttt{compile} groups vector-lifting rewrites, and \texttt{opt} groups both \emph{expose}-style rewrites (e.g., $a \rightsquigarrow a + 0$) and high-profit vector optimizations such as \texttt{vec-opt}. The lower panel shows the corresponding flow. The strategy executes a \texttt{sequence}, in which a \texttt{repeat} node applies \texttt{pre-compile}, \texttt{compile}, and \texttt{simplify} for up to 20 rounds. This repeated block terminates early if no further cost improvement is observed. After that, the strategy applies \texttt{opt} for 10 iterations. 

\section{LLM-Guided Strategy Synthesis}
\label{sec:guided-synthesis}

Once EqSat control is represented as an EqSatL artifact, the remaining challenge is to synthesize effective strategies over that artifact space while keeping search stable under e-graph growth. This section presents \eggmind's offline strategy synthesis methodology, which combines an agentic workflow for iterative strategy search, proof-derived rewrite motif caching for reusable evidence accumulation, and tractability guidance for controlling unstable rewrite interaction.

\subsection{Agentic Workflow}
\label{sec:agent-workflow}
\begin{figure*}[t]
  \centering
  \includegraphics[width=0.88\textwidth]{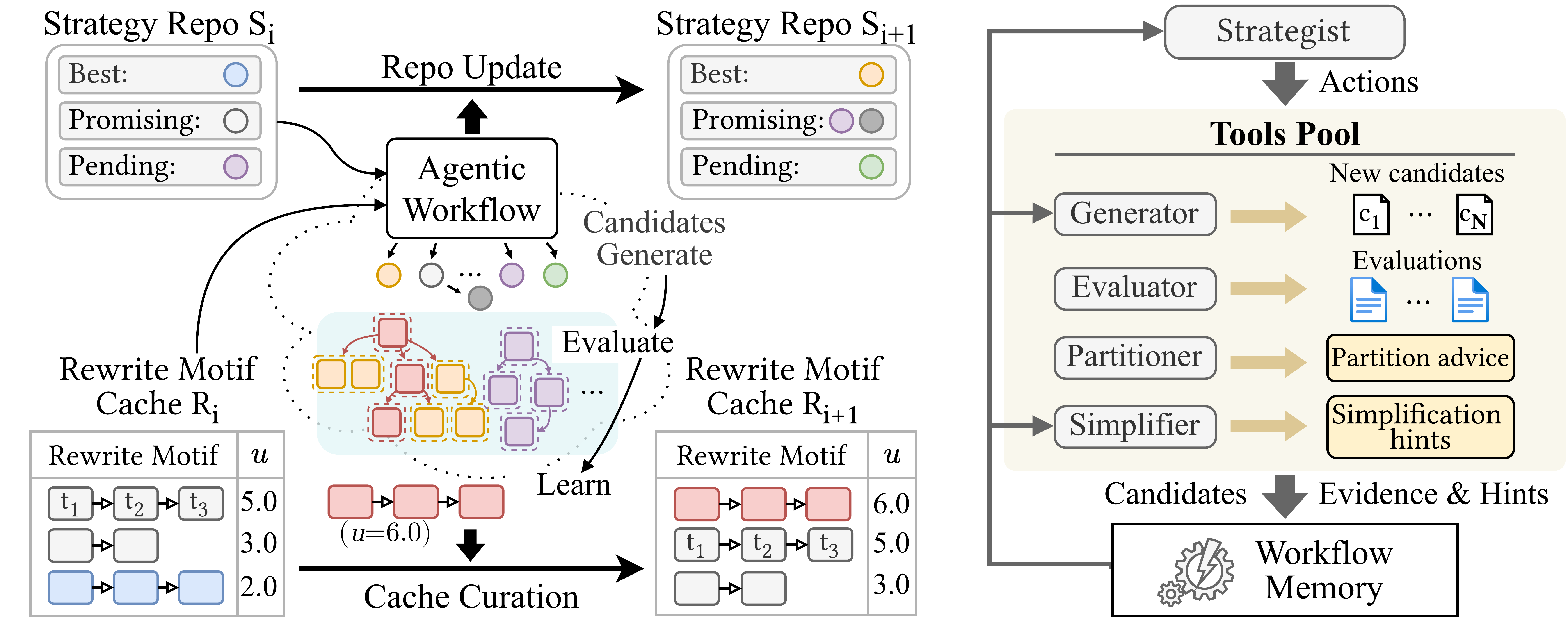}
  \caption{At each iteration of EggMind's offline strategy synthesis, the agentic workflow on the right proposes and evaluates EqSat strategies, updating the candidate pool and rewrite motif cache from iteration $i$ to iteration $i+1$.}
  \label{fig:session-workflow}
  \Description{Operational overview of EggMind's agentic workflow, showing session state, strategist, host, generator, simplifier, evaluation tools, dependency guidance, and proof-derived rewrite motif caching.}
\end{figure*}
\autoref{fig:session-workflow} shows one iteration of \eggmind's offline synthesis process, from iteration $i$ to $i+1$. At iteration $i$, the current search context is summarized by \emph{workflow memory} that consists of a repository of EqSatL strategy artifacts, and a rewrite motif cache (detailed in \autoref{sec:trace-memory}) that stores reusable structural evidence from successful runs. The strategy repository stores three kinds of strategies: the current \textit{best} strategy, a set of \textit{promising} strategies that are worth refining, and \textit{pending} strategies that have been generated but not yet evaluated. Starting from this context, the agentic workflow proposes strategy updates, evaluates with EqSat, and updates the state for the next iteration. In this way, the offline synthesis stage evolves via explicit state updates instead of opaque conversational traces.

Within each iteration, \eggmind organizes strategy search as a controlled decision-execution-feedback loop. The Strategist selects the next action from the current context, such as proposing a new strategy structure, tuning a promising strategy, or requesting additional partition or simplification guidance. The selected action then activates a tool pool containing four specialized agents: the Generator, Evaluator, Partitioner, and Simplifier. 

Specifically, the Generator proposes new strategies, and the Evaluator supports a coarse-to-fine set of budget modes, including full-budget evaluation for final comparison, reduced-budget evaluation for parameter tuning, and minimal-budget structural sanity checks for quickly screening new structures. The Partitioner and Simplifier provide tractability guidance during search by supplying partition advice and simplification hints that reduce the risk of unstable rewrite interaction. Their outputs are written back to workflow memory and reused in later decisions. 
We defer the detailed design of the Partitioner and Simplifier to \autoref{sec:tractability-guidance}.

This workflow is centered on compact and reusable search objects. Strategies serve as the evolving candidates, while execution outcomes provide the evidence for refinement. After evaluation, the best-performing strategy updates the \emph{best} slot, strategies that appear effective but still worth further tuning are retained as \emph{promising}, and generated but as-yet unevaluated strategies remain \emph{pending}. Across iterations, successful runs also contribute reusable information for later rounds. As evidence accumulates across cases during evolution, one strategy is selected for online reuse.

\subsection{Proof-Derived Rewrite Motif Caching}
\label{sec:trace-memory}
\begin{figure}[t]
  \centering
  \includegraphics[width=\columnwidth]{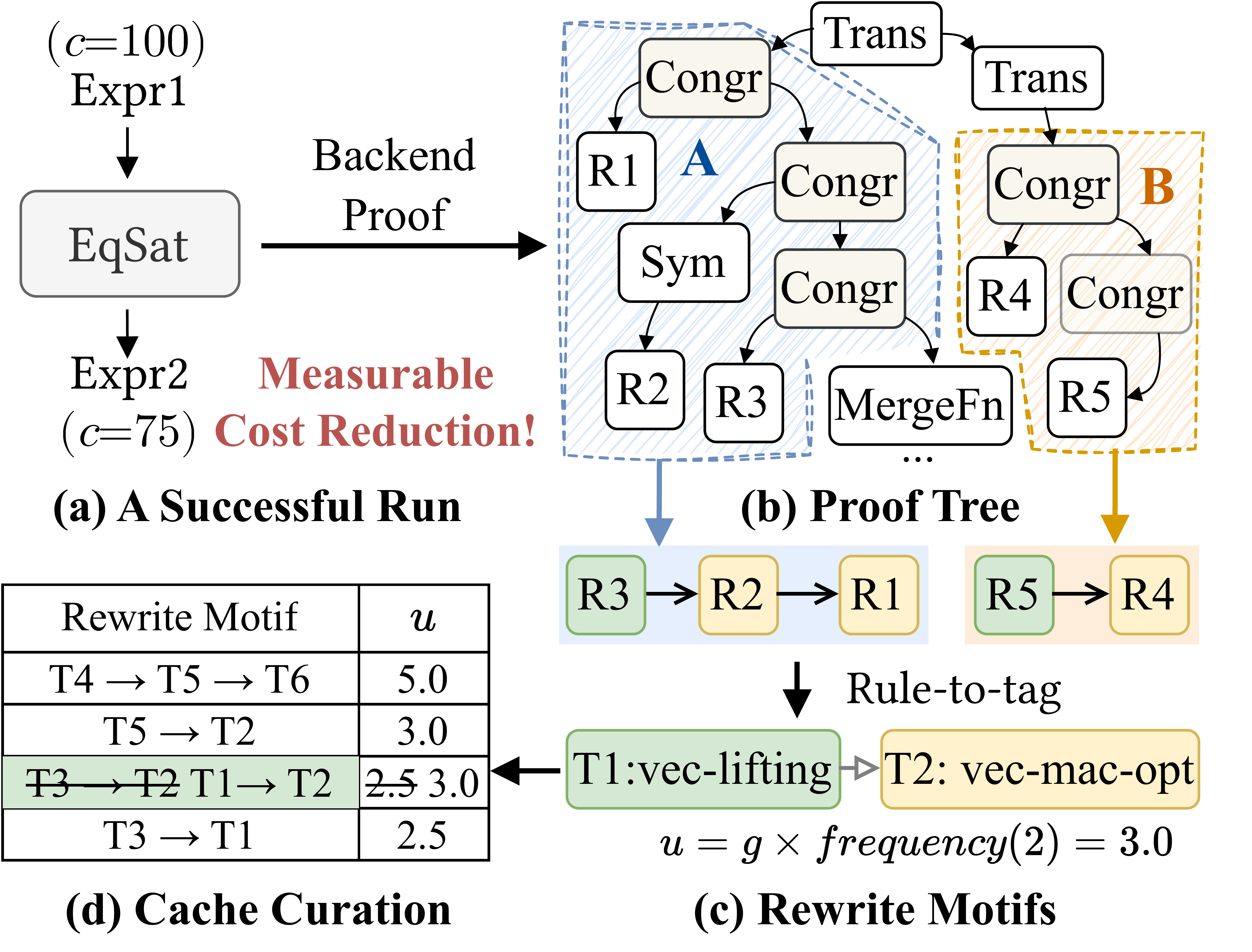}
  \caption{Proof-derived rewrite motif caching in \eggmind. (a) A successful EqSat run. (b) The proof tree, with local regions highlighted for extraction. (c) Rewrite motifs after lifting and deduplication. (d) Utility-based cache curation.}
  \label{fig:proof-memory}
\end{figure}

We next explain how \eggmind\ accumulates proof-derived rewrite motifs from successful strategy evaluations and reuses them to guide later proposals. The key idea is to cache compact rewrite motifs instead of raw proofs. Here, a rewrite motif is a short chain of semantic tags, using the tag vocabulary introduced in \autoref{sec:eqsatl}, that abstracts a concrete rewrite path observed in a successful EqSat run. It captures the trajectory through which an expression evolves toward a better form, while suppressing backend-specific rule details. As shown in \autoref{fig:proof-memory}(a), when the Evaluator executes a candidate strategy and obtains an EqSat run with measurable cost reduction, \eggmind\ requests from the backend a proof of equivalence between the initial and final expressions. \autoref{fig:proof-memory}(b) illustrates the resulting proof tree. Although such proof records justify the improvement, they are not yet a good reusable object for synthesis. They are large, mixed with bookkeeping steps such as congruence and transitivity, and difficult for the LLM to use efficiently in later rounds.

To make this evidence more LLM-friendly, we extract local rewrite motifs from the proof tree. \eggmind\ first identifies local proof regions, shown as the highlighted regions in \autoref{fig:proof-memory}(b). Each region is indexed by a congruence path that locates a local subexpression context in the proof tree, allowing \eggmind\ to isolate a short rewrite chain within that region. This chain is then lifted through a stable rule-to-tag mapping to obtain a higher-level motif. Because multiple backend rules may share the same semantic tag, distinct concrete rewrite chains can collapse to the same higher-level motif. In \autoref{fig:proof-memory}(c), the resulting motif is represented by the tag chain \texttt{vec-lifting} $\rightarrow$ \texttt{vec-mac-opt}, indicating that lifting first exposes the expression to the \texttt{vecmac}-related rewrites that enable further optimization. This lifted and deduplicated motif is then captured by the cache as a smaller and more stable form of proofs.

The cache is updated selectively to retain only high-utility motifs from successful runs. Each motif is scored by combining its observed relative gain, $g = 1 - c_{\mathrm{after}} / c_{\mathrm{before}}$, 
with its frequency across successful cases, and only the strongest entries are kept, as illustrated in \autoref{fig:proof-memory}(d). At reuse time, \eggmind\ exposes only a small set of top-ranked motifs as reusable context for proposing later EqSatL strategies. In this way, the cache serves as a compact source of structural evidence, preserving locally useful rewrite interactions while remaining stable enough for cross-case reuse.

\subsection{Tractability Guidance}
\label{sec:tractability-guidance}
To keep phased EqSat tractable during offline synthesis, \eggmind introduces two forms of tractability guidance: a dependency-based ruleset risk model and LLM-guided simplification hints.

\subsubsection{Dependency-based ruleset risk model}

Large rewrite vocabularies make ruleset partitioning difficult, because poor phase assignments can induce unstable rewrite interaction and rapid e-graph growth. To guide this partitioning problem, \eggmind\ uses a dependency-based ruleset risk model that scores candidate phase assignments and provides partition advice for later strategy proposal.

We model the rewrite vocabulary as a directed rule dependency graph $G=(R,E,w)$, where each node $r\in R$ is a rewrite rule and each weighted edge $w_{ij}$ estimates how strongly rule $r_i$ can enable rule $r_j$ through pattern overlap. A phase assignment $\phi:R\rightarrow\{1,\dots,k\}$ places each rule into one of $k$ phases. We then score a candidate partition with the following objective:

{
\footnotesize
\begin{align}
\max_{\phi}\;&
\alpha \sum\nolimits_{(i,j)\in E} w_{ij}\,\mathbf{1}[\phi(i)+1=\phi(j)]
&& \text{(forward)} \nonumber\\
&-\beta \sum\nolimits_{(i,j)\in E} w_{ij}\,\mathbf{1}[\phi(i)>\phi(j)]
&& \text{(backflow)} \nonumber\\
&-\gamma \sum\nolimits_{(i,j)\in E} w_{ij}\,\mathbf{1}[\phi(j)=1,\phi(i)\neq 1]
&& \text{(phase-1 inflow)} \nonumber\\
&-\delta \sum\nolimits_{(i,j)\in E} w_{ij}\,\mathbf{1}[\phi(i)=\phi(j)]
&& \text{(same-phase)}.
\end{align}
}

Here, $w_{ij}$ combines three signals of potential rewrite interaction: direct enablement when the RHS of $r_i$ matches the LHS of $r_j$, operator/argument overlap between the two rules, and subtree-level matchability when the RHS of $r_i$ can match a subtree of $r_j$. The coefficients $\alpha,\beta,\gamma,\delta$ are fixed per domain by a lightweight grid sweep over the cases used for offline strategy evolution.
The objective promotes \emph{forward} flow by rewarding dependencies that move from phase $t$ to phase $t+1$. To maintain search stability, it also penalizes patterns that lead to e-graph instability. \emph{Backflow} is discouraged so that later phases do not reactivate earlier ones, and \emph{phase-1 inflow} is limited to keep the first phase locally coherent. The objective also penalizes \emph{same-phase} entanglement, encouraging strongly interacting rules to be partitioned rather than left collapsed in a single block. As a result, the induced schedules favor forward activation while reducing uncontrolled search expansion.

\subsubsection{LLM-Guided Simplification Hints}


The second form of tractability guidance is phase-aware simplification. Once the EqSat search is organized into phases, simplification becomes an explicit control surface that must balance e-graph contraction with the preservation of intermediate forms that may still matter later. This makes the simplification phase-local. At phase $t$, the system must decide which representations are redundant in the current context and which should be retained for later phases. Heuristic simplification can contract the e-graph locally, but it does not capture which forms are likely to remain useful downstream. \eggmind\ addresses this with an LLM-guided Simplifier that provides phase-specific structural preferences.

For each phase $t$, the Simplifier proposes a small set of \emph{preferred/pruned pattern pairs} $(p^{+},p^{-})$. Together, these pairs define a phase-specific preference order over patterns, which we refer to as a \emph{simplification poset}. We write $p^{-}\prec_t p^{+}$ when, within phase $t$, an e-class that already contains a representative matching $p^{+}$ should treat representatives matching $p^{-}$ as lower-priority candidates. In effect, the Simplifier marks forms that need not be carried forward when a better alternative is already available in the same e-class. This makes simplification depend on phase-local structural preferences, not on a single global notion of what is cheapest or simplest.



\eggmind\ incorporates these hints as additive penalties rather than hard deletions. The simplification score is $s(e)=c(e)+p_{\mathrm{llm}}(e)$, where $c(e)$ is the base cost assigned by the simplification cost model and $p_{\mathrm{llm}}(e)$ is a bounded penalty applied to disfavored patterns. Within each e-class, enodes are ranked by $s(e)$, and the top $\theta$ fraction is pruned. Thus, $p_{\mathrm{llm}}(e)$ biases simplification toward phase-preferred forms, while $\theta$ controls pruning aggressiveness. This keeps the simplification phase-aware without deterministically removing every disfavored candidate. If a set of hints proves too aggressive or harms downstream optimization, later evaluation can reveal the problem, allowing the Simplifier to revise its preferences in subsequent rounds.

\section{Evaluation}
\label{sec:eval}
\begin{figure*}[t]
  \centering
  \includegraphics[width=\linewidth]{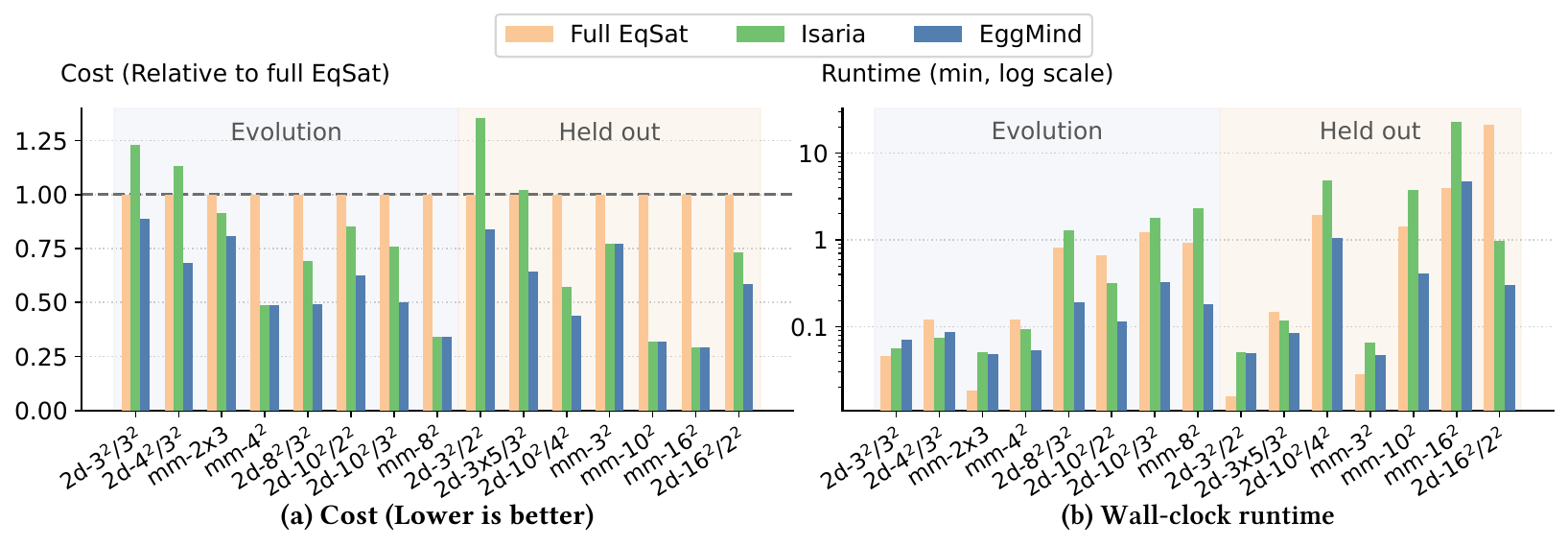}
  \caption{Comparison of the vectorization benchmarks. The x-axis lists representative evolution and test cases.}
  \label{fig:isaria-main}
\end{figure*}

We evaluate \eggmind by addressing the following questions:
\begin{enumerate}[leftmargin=*]
  \item Does \eggmind\ improve the practical resource-quality trade-off over full EqSat and expert-tuned baselines on vectorization benchmarks? (\autoref{sec:isaria-main-results})
  \item How much do \eggmind’s core components contribute to its overall performance? (\autoref{sec:isaria-ablation})
  \item Can \eggmind's methodology be applied effectively across different optimization domains? (\autoref{sec:xla-transfer})
  \item Can \eggmind\ support more automated and reusable e-graph optimizer construction in augmented rewrite spaces? (\autoref{sec:eqmap-case-study})
\end{enumerate}

\paragraph{Methodology.}
For each case, we report the final cost, as defined by each benchmark's cost function, wall-clock runtime, and peak memory usage during EqSat execution. All extraction uses the default greedy extractor in \texttt{egglog}. For offline synthesis, we additionally report synthesis wall-clock time, the number of model requests, and input/output token usage. By default, we use Doubao-Seed-2.0-pro as the base LLM model during offline synthesis. 
All experiments are conducted on a machine with two Intel Xeon Gold 6348 CPUs (56 physical cores) and 2 TiB of RAM. All EqSat execution is performed on CPU. Each online run is subject to a 600\,s wall-clock timeout and a 25\,GB memory limit.

\subsection{Vectorizing Compiler}
\label{sec:vectorization}

We use the same vectorization benchmark family as Isaria and Diospyros~\citep{thomas_automatic_2024,vanhattum_vectorization_2021}, covering both 2D-convolution~($2\mathrm{D}$) and matrix-multiplication~($\mathrm{MM}$) workloads. We use 8 cases for evolution, and use the remaining for held-out testing.

\begin{figure*}[t]
  \centering
  \includegraphics[width=\linewidth]{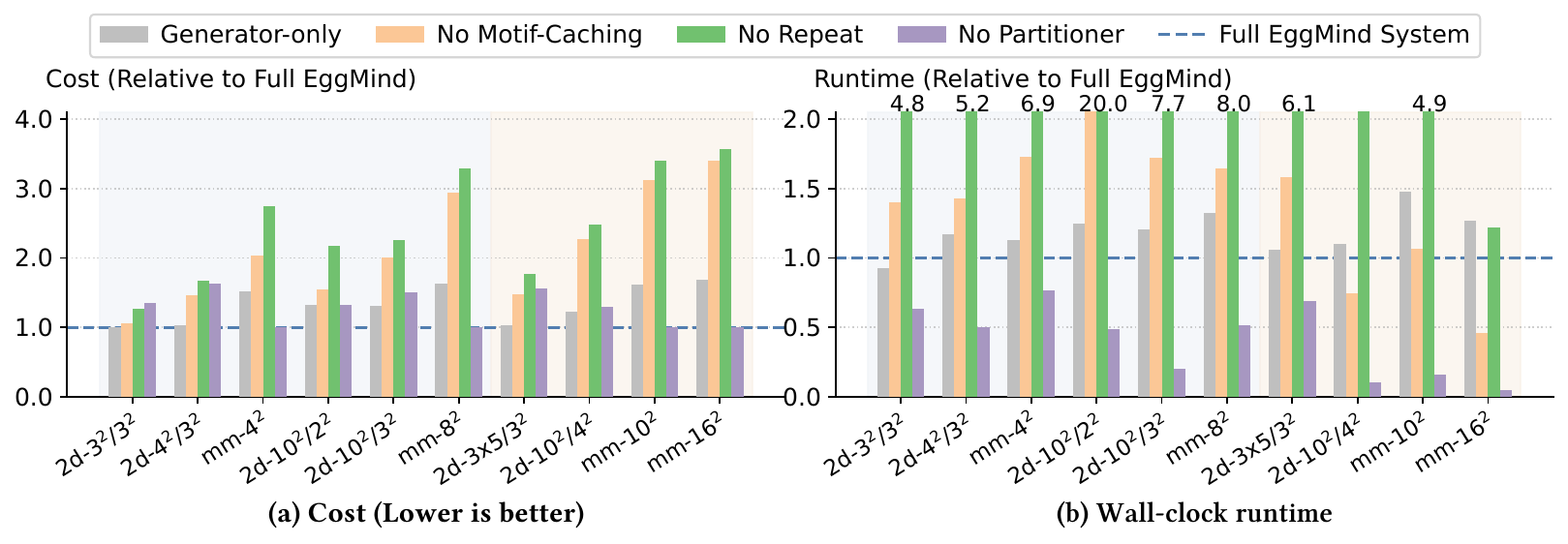}
  \caption{Ratios in the vectorization component-attribution study on representative cases. The x-axis lists representative evolution and test cases. Runtime ratios above 2.0 are truncated for readability and annotated with their exact values.}
  \label{fig:isaria-ablation}
\end{figure*}

\subsubsection{End-to-End Effectiveness}
\label{sec:isaria-main-results}

\autoref{fig:isaria-main} compares \eggmind\ against both unguided (full) EqSat and the expert-tuned Isaria strategy on the main vectorization benchmark family. On this benchmark, \eggmind\ improves the resource--quality trade-off over both baselines.


Optimization quality is measured by final extraction cost under a latency-oriented cost model. \autoref{fig:isaria-main}(a) shows that \eggmind\ reduces cost by a geometric mean of 45.1\% relative to full EqSat and 20.6\% relative to Isaria's strategy. It matches or outperforms all baselines across the evaluated cases, with strict improvements in 10 cases and a maximum gain of 39.9\%. By contrast, Isaria's schedule is occasionally outpaced by full EqSat on small kernels such as $2\mathrm{D}\ 3^2 \times 3^2$, suggesting that fixed hand-crafted schedules do not always preserve the most useful search opportunities.


For resource usage, \eggmind\ also achieves substantial reductions relative to the other baselines. As shown in \autoref{fig:isaria-main}(b), \eggmind\ achieves a geometric-mean online runtime speedup of 2.21$\times$ over Isaria, with per-case speedups reaching 6.85$\times$ over full EqSat on large cases. It is also faster than full EqSat across the evaluated cases, whereas Isaria is not uniformly so. On $2\mathrm{D}\ 16^2 \times 2^2$, for example, \eggmind\ reaches a solution in 18.0\,s and achieves a final cost 41.6\% lower than full EqSat. By comparison, full EqSat and Isaria's strategy require 123\,s and 58\,s, respectively, yet still do not match \eggmind's cost. Peak-memory behavior follows the same trend. Overall, \eggmind\ reduces peak RAM usage by 69.1\% on a geometric-mean basis relative to full EqSat, with the largest reductions exceeding 95\%, for example from 21.6\,GB to 0.74\,GB on $\mathrm{MM}\ 20^2 \times 20^2$. Compared with Isaria under the same setup, \eggmind\ also reduces peak memory by 11.4\% on average across comparable cases, with the largest reduction reaching 50.4\% on $\mathrm{MM}\ 8^2 \times 8^2$.

A qualitative inspection of the synthesized strategies helps explain this behavior. \eggmind\ tends to place expansive \emph{expose} rewrites, such as $a \to a+0$, inside \texttt{repeat} regions together with vector lifting, local vector normalization, and simplification, instead of letting them interact freely with later optimization phases. By contrast, the expert-designed Isaria schedule shown in \autoref{fig:eqsatL-example} places several high-risk rewrites inside a late-stage \texttt{opt} phase with a large iteration count, so repeated activation amplifies e-graph growth without a comparable gain in search quality. This lets \eggmind\ retain their optimization value while mitigating uncontrolled search expansion.

Beyond performance, \eggmind\ achieves these results with only moderate offline synthesis overhead. Offline synthesis over the vectorization benchmark family takes 31.8 minutes and requires 53 model requests, using 4.35M input tokens and 36.6K output tokens, for an estimated cost of roughly \$3.16 per synthesis run under current pricing. This is a one-time cost for each workload family and can be amortized over later online runs.

\subsubsection{Ablation Study.} 
\label{sec:isaria-ablation}


To isolate the contribution of each major design choice in \eggmind, we conduct an ablation study on the vectorization benchmark, comparing the full system against four variants: (1) Remove strategist (\textit{Generator-only}), (2) Remove motif reuse (\textit{No Motif-Caching}), (3) Disable \texttt{repeat} construct in EqSatL (\textit{No Repeat}), and (4) Bypassing the dependency-based ruleset risk model(\textit{No Partitioner}). \autoref{fig:isaria-ablation} reports the final cost and online EqSat runtime relative to the full \eggmind\ system. The following statistics are derived from the same 15 cases used in the main comparison, while the figure visualizes 10 of them for readability.

All four components in \eggmind\ contribute to the overall performance, with the largest degradations occurring when motif caching or \texttt{repeat} is removed. Without motif caching, cost worsens by a geometric mean of 78.3\%, while runtime and peak memory increase by 53.0\% and 25.3\%, respectively, consistently yielding the highest costs. On $2\mathrm{D}\ 10^2 \times 2^2$, runtime increases from 6.94\,s to 25.35\,s and cost rises from 3305 to 5114. 
Removing \texttt{repeat} has an even larger effect. On average, the cost worsens by 118.7\%, while the runtime and peak memory further increase by 475.6\% and 91.8\%, respectively, and all cases become both slower and worse. On $\mathrm{MM}\ 16^2 \times 16^2$, cost worsens by 3.57$\times$ and memory increases from 639.4\,MB to 2.5\,GB. These results show that motif reuse and repeated multi-phase interaction are both critical for search quality and stability.
The strategist and rule partitioner play distinct roles. Removing the strategist degrades cost by 24.5\% and increases runtime by 16.3\%, with cost worsening in all cases, indicating benefits beyond one-shot generation. Removing the rule partitioner worsens cost by 23.1\% but reduces runtime by 63.3\%, with all cases faster and 10/15 worse in cost. This suggests that partitioning mainly improves optimization quality by organizing interacting rewrites into effective schedules, and lower search cost is a secondary effect. Although the full system incurs higher runtime, it achieves better solutions in most cases, making rule partitioning worthwhile when prioritizing quality. Overall, \eggmind's gains arise from the combination of strategist-guided refinement, motif reuse, repeated interaction, and explicit rewrite partitioning, suggesting a strong synergy between our proposed techniques.

\paragraph{Role of Simplification.}
\begin{figure}[t]
  \centering
  \includegraphics[width=.9\columnwidth]{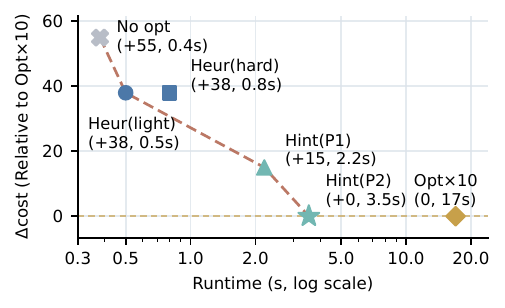}
  \caption{Trade-off among simplification controls.}
  \label{fig:simplifier-tradeoff}
  \Description{A single-column trade-off plot for the representative Isaria 2D convolution case 10 squared by 2 squared. The horizontal axis is runtime in seconds on a log scale and the vertical axis is final-cost delta relative to the original opt times 10 baseline. The plotted points compare no opt, heuristic simplification with two strength settings, Poset1, Poset2, and the original opt times 10 baseline, showing that phase-local control can reduce runtime substantially while keeping final cost close to the baseline.}
\end{figure}

On the vectorization benchmark, simplification is a core component of successful strategies. Both Isaria and \eggmind\ place explicit \texttt{simplify} nodes at phase boundaries and inside refinement loops, reflecting the need to control e-graph growth while preserving useful intermediate forms. 
For most workloads, \eggmind's ruleset partitioning and schedule construction already steer e-graph growth in useful directions. Heuristic simplification therefore does not need to identify which forms remain valuable later and can mainly focus on contraction, so disabling hinted simplification changes little.

However, simplification acts as a crucial safeguard when rulesets are arbitrary and disjoint. To evaluate the effectiveness of hints more directly, we study a targeted high-growth setting derived from Isaria's \texttt{opt}$\times 10$ block(\autoref{fig:eqsatL-example}), where \emph{expose}-style and vector-optimization rewrites are mixed in one coarse phase. In this setting, heuristic simplification can become myopic because it contracts locally without recognizing which intermediate forms still matter for later progress. \autoref{fig:simplifier-tradeoff} compares simplification controls under a fixed iteration budget. Skipping \texttt{opt} eliminates runtime (0.4\,s) but incurs a +55 cost delta. \emph{Heuristic simplification} with varying strengths already retains most of the benefit, keeping results closer to baseline (+38) while reducing runtime to 0.5-0.8\,s, where \textit{light} and \textit{hard} correspond to the strength parameter $\theta$. We further evaluate \emph{hint-guided simplification} through two preference posets proposed by the Simplifier. \textit{Poset1} orders commutative variants and reaches +15 in 2.2\,s. \textit{Poset2} extends it with a \texttt{vecmac}-specific preference and recovers baseline quality in 3.54\,s. These results show that simplification itself is essential, and that hints are most useful when coarse phase structure makes heuristic contraction less reliable.

\paragraph{Necessity of EqSatL and Specialized Agentic Flow.}
\label{sec:isaria-search-dynamics}
\begin{table}[t]
  \centering
  \footnotesize
  \caption{Comparison against free-agent search.}
  \label{tab:free-agent-stress}
  \begin{tabular}{llccc}
    \toprule
    \textbf{Method} & \textbf{Agent} & \textbf{Time} & \textbf{\#EqSat-runs} & \textbf{Gain} \\
    \midrule
    \multirow{4}*{\makecell{Free-search \\ over \textit{egglog}}} & Kimi Code (K2.5) & 33.0 min & 42  & +0.0\% \\
    & Codex (GPT-5.4)  & 47.0 min & 22  & +0.0\% \\
    & Codex (GPT-5.3)  & 46.0 min & 46  & +0.0\% \\
    & ClaudeCode (DB)  & 145.0 min & 12  & +0.0\% \\
    \midrule
    \multirow{4}*{\makecell{Free-search \\ over \textit{EqSatL}}} & Kimi Code (K2.5) & 37.0 min & 160  & +15.1\% \\
    & Codex (GPT-5.4)  & 51.8 min & 103  & +0.0\% \\
    & Codex (GPT-5.3)  & 44.5 min & 208  & +18.6\% \\
    & ClaudeCode (DB)  & 20.0 min & 16  & +0.0\% \\
    \midrule
    \eggmind         & — & 31.8 min & 80  & +47.8\% \\
    \bottomrule
  \end{tabular}
\end{table}

We compare \eggmind with several commercial agent systems on the vectorization benchmark. These agents include: Kimi Codes (with Kimi K2.5), Codex (with GPT 5.3 and 5.4 High), and Claude Code (Doubao-Seed-2.0-pro, same as \eggmind's). We ask these agent systems to search for EqSat strategies by generating egglog scripts or EqSatL strategies. The results are shown in \autoref{tab:free-agent-stress}.

We first demonstrate why direct LLM-driven evolution over raw egglog is difficult in practice. 
All four agents operating on raw egglog only achieve parity with the full EqSat baseline on average. Under comparable search time, they complete far fewer EqSat runs than those searching over EqSatL. 
This suggests that much of the budget is consumed by constructing and repairing low-level scripts, leaving fewer chances to evaluate and refine strategy candidates. Specifically, although Claude Code is equipped with the same base model as \eggmind, it completes only 12 raw-egglog EqSat runs in 145.0 minutes and still achieves no held-out gain. 

Beyond EqSatL, we then demonstrate why \eggmind's specialized workflow is also essential in this task. 
Free-agent performance over EqSatL remains highly variable, ranging from +0.0\% to +18.6\% mean held-out gain. By contrast, \eggmind\ reaches 47.8\% mean held-out cost gain while using fewer EqSat runs than the strongest EqSatL-based free agent. This indicates that EqSatL provides the right abstraction boundary, while the bounded workflow is what makes search reliably effective and sample-efficient.



These results directly reflect the challenges described in \autoref{sec:code-evolution}. Searching over raw egglog lacks a reusable strategy abstraction, provides weak feedback for refinement, and makes tractability much harder to control. EqSatL addresses the abstraction problem, while \eggmind's specialized workflow further turns this into a reliable and efficient search.


\subsection{XLA-Based Tensor Compiler}
\label{sec:xla-transfer}

To evaluate \eggmind\ beyond the vectorization domain, we turn to tensor graph optimization. Initial experiments on standard workloads from prior work, such as Tensat~\citep{yang2021equalitysaturationtensorgraph} and \textsc{rcmcts}~\citep{hartmann_tensor_eqsat_mcts_2024}, showed limited room for strategy synthesis: in many cases, unguided EqSat already reached its best cost within one or two iterations. We therefore focus on the algebraic-simplification pass of XLA HLO~\citep{xla_tf2xla,arora_tensorright_2025} and construct a more demanding benchmark by extending the rewrite space. 

In this XLA-based setting, \eggmind\ continues to improve the resource--quality trade-off. Across 17 evaluation cases, the synthesized strategies achieve lower final cost than the unguided baseline in 13 cases and match it in the remaining 4, while reducing online EqSat runtime by a geometric mean of 11.89$\times$ relative to full EqSat. These results show that \eggmind's offline synthesis methodology transfers beyond the original vectorization benchmark, especially in settings where unguided exploration remains slow and unstable under an enlarged rewrite space.

\subsection{Case Study on Logic Synthesis}
\label{sec:eqmap-case-study}

\begin{figure}[t]
  \centering
  \includegraphics[width=\columnwidth]{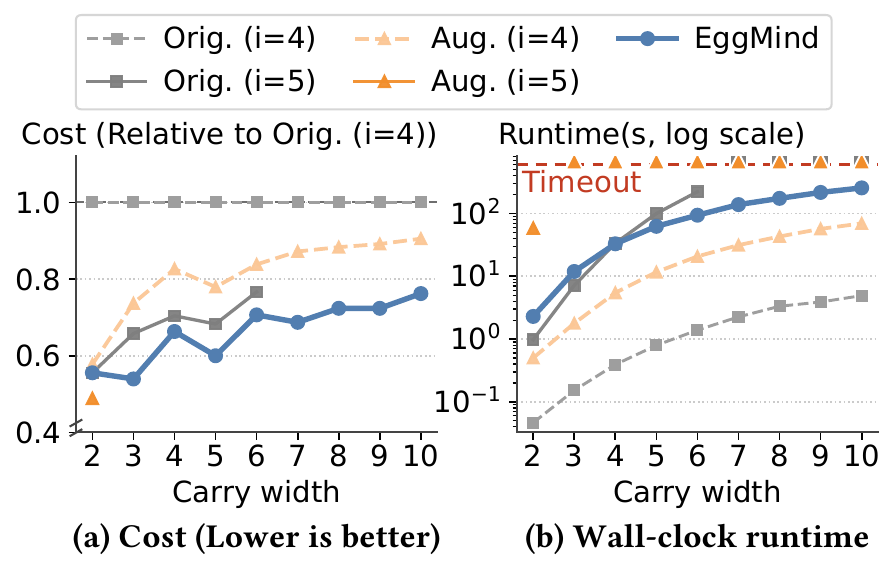}
  \caption{EqMap carry-benchmark comparison for cost reduction and wall-clock runtime across five configurations.}
  \label{fig:eqmap-main-comparison}
  \Description{A single-column EqMap comparison figure on carry benchmarks of increasing width, with two side-by-side panels that share the x-axis. Both panels use carry-chain width on the x-axis. Panel (a), labeled Quality, shows the final cost normalized to the Original Rules with iteration 4; smaller is better. Panel (b), labeled Runtime, shows wall-clock runtime in seconds on a log scale, with a timeout reference line at 600 seconds; smaller is better. The compared configurations are Original Rules with iteration 4, Original Rules with iteration 5, Augmented Rules with iteration 4, Augmented Rules with iteration 5, and EggMind's synthesized strategy.}
\end{figure}

In this case study, we examine \eggmind with technology mapping in logic synthesis. Our rewrite space, derived from the ASIC EqMap flow~\citep{hofmann_eqmap_2025}, combines Boolean equivalence rewrites for logical exploration with lifting rules for technology mapping, and is evaluated on carry-chain benchmarks of increasing bit-widths. While expressive rewrites and large budgets can, in principle, expose lower-cost cell mappings, many profitable mappings require long rewrite chains. Under the original EqMap flow, such mappings typically emerge only with large global iteration budgets, which become computationally prohibitive as circuit width grows. The key bottleneck is therefore not expressiveness, but practical reachability under bounded search.

\paragraph{Rule Augmentation with Enumo.}
We use Enumo~\citep{pal_equality_2023} to synthesize interaction rewrites between Boolean logic and physical cells. 
Two representative rules are shown below:

\begingroup
\small
\[
\begin{aligned}
\mathsf{Or}\bigl(
  &\mathsf{AOI21}_{\mathsf{X1}}(c,b,a),\;
   \mathsf{AOI21}_{\mathsf{X1}}(d,b,a)
\bigr)
  \Longrightarrow
  \mathsf{AOI22}_{\mathsf{X1}}(d,c,b,a), \\
\mathsf{And}\bigl(
  &\mathsf{OAI21}_{\mathsf{X1}}(d,b,a),\;
   \mathsf{OAI21}_{\mathsf{X1}}(c,b,a)
\bigr)
  \Longrightarrow
  \mathsf{OAI22}_{\mathsf{X1}}(b,a,c,d).
\end{aligned}
\]
\endgroup

These rewrites collapse long chains of Boolean normalization, factoring, and lifting into single-step transformations, enabling profitable higher-level cell mappings to be reached earlier in the search. However, they also enlarge the matching space and make the search more susceptible to e-graph explosion. Rule augmentation, therefore, improves reachability while making the expanded rewrite space harder to exploit in practice. \eggmind\ addresses this with schedule construction and simplification in EqSatL.

\paragraph{End-to-End Results.}

We compare \eggmind against the original and augmented rulesets evaluated at two distinct EqSat iterations($i=4$ and $i=5$), where \eggmind operates over the same augmented rewrite space. This allows us to differentiate  the gains from search budget, rewrite-space reachability, and strategic organization.  
\autoref{fig:eqmap-main-comparison} shows that \eggmind\ achieves the best quality-efficiency trade-off. Increasing the budget for the original ruleset ($i=4\!\to\!5$) improves cost but causes four large cases to exceed the 600\,s timeout. Augmented Rules at $i=4$ improve reachability, but at $i=5$ expose the inherent limits of unguided search. The e-graph expansion here becomes so explosive that only the smallest test cases finish within 600\,s, and reach $\sim$4\,GB memory usage. Thus, rule augmentation alone is insufficient, and enlarging the matching space makes brute-force search even less attractive.  
By contrast, \eggmind\ converts the same augmented space into a stable scheduled optimization process, achieving the lowest cost on 8/9 benchmarks (with unguided Augmented Rules ($i=5$) winning only on the smallest case), reducing cost by 33.76\% and 8.75\% compared to Original Rules with $i=4$ and $i=5$, respectively, while maintaining stable runtime (2.3-255.4\,s). It also reduces peak memory by 51.94\% compared to the unguided $i=5$. These results underscore the synergy between rule and strategy synthesis: the former expands reachability, while the latter determines whether that expanded space is practically exploitable.

\section{Related Work}
\label{sec:related}

Modern equality saturation builds on efficient e-graph substrates such as \texttt{egg}~\citep{max_willsey_egg_2021} and unified datalog frameworks such as \texttt{egglog}~\citep{zhang_better_2023}. Other work improves EqSat scalability through better extraction, domain-specific pruning, or persistent e-graph integration into compiler IRs~\citep{cai2025smoothe,yin2025boost,emorphic_25_chen,eqsatdialect}. Our focus is different: we study how EqSat search itself is partitioned, scheduled, and controlled.

Systems like \ruler~\citep{nandi_rewrite_2021} and \enumo~\citep{pal_equality_2023} automate rewrite discovery, making rewrite spaces easier to construct but harder to manage. More broadly, strategic rewriting has long separated rewrite rules from the strategies that govern where, when, and in what order they are applied, as in Stratego~\citep{visser_2001_stratego}, ELAN~\citep{borovansky_2001_elan}, and Maude~\citep{maude_strategy_language}. For EqSat, prior work has shown the value of such control through expert-designed phased schedules in Isaria~\citep{thomas_automatic_2024}, guide-based steering with intermediate targets~\citep{koehler_sketch_guided_eqsat_2022,koehler_guided_eqsat_2024}, and customizable saturation loops with multi-phase rewriting at the engine level~\citep{parallel-eqsat}. \eggmind\ builds on this line of work, but targets offline synthesis of EqSat strategy artifacts reusable across related cases.

Recent work has also applied LLMs to compiler optimization and e-graph control, including compiler-oriented foundation models, open-ended code evolution, and online optimization controllers~\citep{cummins2024metalargelanguagemodel,novikov2025alphaevolvecodingagentscientific,chen2026magellanautonomousdiscoverynovel,qiu2026passbypassoptimizationintentdrivenir,mikek2026agenticcodeoptimizationcompilerllm,Zhang2025ASPENLE}. Unlike online controllers or free-form code evolution, \eggmind\ synthesizes explicit EqSat strategy artifacts over related cases, with control surfaces for ruleset partitioning, schedule construction, and simplification. This makes strategy-discovery cost reusable and amortizable across future runs.

\section{Conclusion}
\label{sec:conclusion}

\eggmind reframes equality saturation strategy design as offline synthesis over the explicit, reusable EqSatL DSL. It enables LLM-guided synthesis of strategies that can be validated, reused, and amortized across problem instances of the same domain. Through proof-derived motif reuse and tractability guidance, \eggmind makes EqSat more effective. Overall, it provides a practical path toward automating the strategy layer of modern e-graph-based compilers.

\newpage

\bibliographystyle{ACM-Reference-Format}
\bibliography{refs}

\end{document}